\pdfoutput=1
\documentclass{article}
\usepackage{etex}

 \usepackage[nonatbib,final]{nips_2016}

\usepackage[utf8]{inputenc} 
\usepackage[T1]{fontenc}    
\usepackage{url}            
\usepackage{booktabs}       
\usepackage{amsfonts}       
\usepackage{nicefrac}       
\usepackage{microtype}      

\usepackage{url}
\usepackage{graphicx}
\usepackage{soul}
\usepackage{color}
\usepackage{epsfig}
\usepackage[belowskip=0pt,aboveskip=0pt,font=small]{caption}
\usepackage[belowskip=0pt,aboveskip=0pt,font=small, position=t]{subcaption}
\usepackage{amsmath, amsthm, amssymb}
\usepackage{textcomp}
\usepackage{stmaryrd}
\usepackage{upgreek}
\usepackage{bm}
\usepackage{cases}
\usepackage{mathtools}
\usepackage[pagebackref=false,breaklinks=true,letterpaper=true,colorlinks,bookmarks=false,citecolor=red]{hyperref}
\usepackage[ruled,noend]{algorithm2e}
\usepackage{algpseudocode}
\usepackage{multirow}
\usepackage{rotating}
\usepackage{booktabs}
\usepackage{enumitem}
\usepackage[olditem,oldenum]{paralist}
\usepackage{alltt}
\usepackage{listings}
\usepackage{mysymbols}
\usepackage{xspace}
\usepackage{comment}
\usepackage{afterpage}
\usepackage{pdfpages}
\usepackage{framed}
\usepackage{pgfplots}
\usepackage{setspace}
\usepackage{array}
\usepackage{cite}
\pgfplotsset{
  every tick label/.append style={font=\tiny},
  every axis plot/.append style={line width=1.2pt},
  every axis title/.append style={scale=0.9,yshift=-0.85em},
  every axis label/.append style={scale=0.75},
  every axis plot post/.append style={
    every mark/.append style={line width=1pt}
  }
}

\belowdisplayskip \abovedisplayskip

\newlength{\sectionReduceTop}
\newlength{\sectionReduceBot}
\newlength{\subsectionReduceTop}
\newlength{\subsectionReduceBot}
\newlength{\abstractReduceTop}
\newlength{\abstractReduceBot}
\newlength{\captionReduceTop}
\newlength{\captionReduceBot}
\newlength{\subsubsectionReduceTop}
\newlength{\subsubsectionReduceBot}

\newlength{\eqnReduceTop}
\newlength{\eqnReduceBot}

\newlength{\horSkip}
\newlength{\verSkip}

\newlength{\figureHeight}
\title{Stochastic Multiple Choice Learning for \\Training Diverse Deep Ensembles}

\definecolor{lightblue}{rgb}{.60,.75,1}

\newcommand{\xhdr}[1]{{\vspace{2pt}\noindent\textbf{\textit{#1}}}}

\author{%
Stefan Lee\\Virginia Tech\\steflee@vt.edu%
\And Senthil Purushwalkam \\Carnegie Mellon University\\spurushw@andrew.cmu.edu\And%
Michael Cogswell\\Virginia Tech\\cogswell@vt.edu \And%
Viresh Ranjan\\Virginia Tech\\rviresh@vt.edu%
\AND  David Crandall\\Indiana University\\ djcran@indiana.edu%
\And  Dhruv Batra\\Virginia Tech\\dbatra@vt.edu}

\begin{document}

\maketitle

\begin{abstract}
Many practical perception systems exist within larger processes that include interactions with users or additional components capable of evaluating the quality of predicted solutions. In these contexts, it is beneficial to provide these oracle mechanisms with multiple highly likely hypotheses rather than a single prediction. In this work, we pose the task of producing multiple outputs as a learning problem over an ensemble of deep networks -- introducing a novel stochastic gradient descent based approach to minimize the loss with respect to an oracle. Our method is simple to implement, agnostic to both architecture and loss function, and parameter-free. Our approach achieves lower oracle error compared to existing methods on a wide range of tasks and deep architectures. We also show qualitatively that the diverse solutions produced often provide interpretable representations of task ambiguity.
\end{abstract}

\section{Introduction}
\label{sec:intro}

Perception problems rarely exist in a vacuum. Typically, problems in Computer Vision, 
Natural Language Processing, and other AI subfields are embedded in larger applications 
and contexts. For instance, the task of recognizing and segmenting objects in an image (semantic 
segmentation \cite{pascal-voc-2011}) might be embedded in an autonomous vehicle \cite{geiger_ijrr}, while the 
task of describing an image with a sentence (image captioning \cite{coco}) might be part of a
system to assist visually-impaired users \cite{msr-blindcaption,fb-blindcaption}.  

In these scenarios, the goal of perception is often not to generate a single output but \emph{a set of plausible 
hypotheses} for a `downstream' process, such as a verification component or a human operator. 
These downstream mechanisms may be abstracted as \emph{oracles} that have the capability to pick the correct 
solution from this set. Such a learning setting is called Multiple Choice Learning (MCL) \cite{rivera_nips12}, where 
the goal for the learner is to minimize oracle loss achieved by a set of $M$ solutions. 
More formally, given a dataset of input-output pairs $\{(x_i,y_i)\mbox{ | }x_i\in \mathcal{X}, y_i\in \mathcal{Y}\}$, the goal of classical supervised learning is to search for a mapping $F:\mathcal{X} \rightarrow \mathcal{Y}$ that minimizes a task-dependent loss $\ell: \mathcal{Y}\times \mathcal{Y} \rightarrow \mathcal{R}^+$ 
capturing the error between the actual labeling $y_i$ and predicted labeling $\hat{y}_i$. In this setting, the 
learned function $f$ makes a single prediction for each input and pays a penalty for that prediction. In contrast, 
Multiple Choice Learning seeks to learn a mapping $g: \mathcal{X} \rightarrow \mathcal{Y}^M$ that 
produces $M$ solutions $\hat{Y}_i = (\hat{y}^1_i,\ldots,\hat{y}^M_i)$ such that oracle loss $\min_{m} \ell\left(y_i,\hat{y}^m_i\right)$ is minimized.

In this work, we fix the form of this mapping $g$ to be the 
union of outputs from an ensemble of predictors such that $g(x) = \{f_1(x), f_2(x), \dots, f_M(x)\}$, 
and address the task of training ensemble members $f_1,\dots,f_M$ such that $g$ minimizes oracle loss.
Under our formulation, different ensemble members are free to specialize on subsets of the data
distribution, so that collectively  they produce a set of outputs which covers the space of high probability predictions well.

\begin{figure}[t]
\captionsetup[subfigure]{aboveskip=2pt, belowskip=-1pt}
\begin{subfigure}[t]{0.4\columnwidth}
\centering
\resizebox{\columnwidth}{0.75in}{\includegraphics{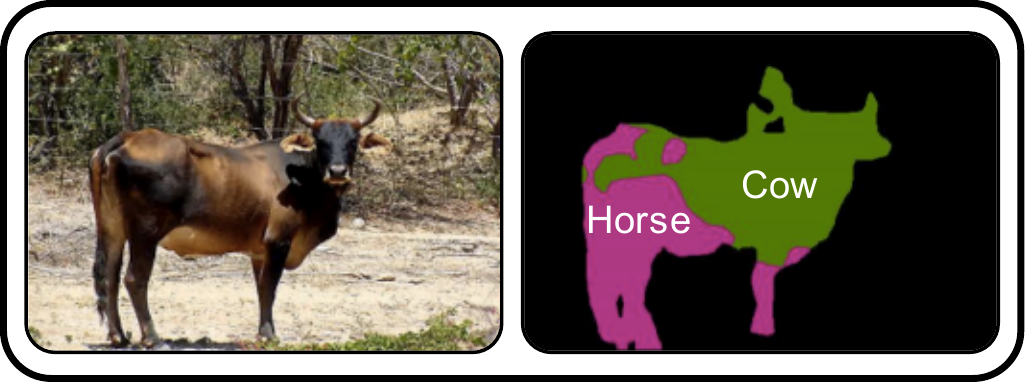}}
\label{fig:mixedseg}
\end{subfigure}\vspace{-8pt}
\begin{subfigure}[t]{0.6\columnwidth}
\centering
\resizebox{\columnwidth}{0.75in}{\includegraphics{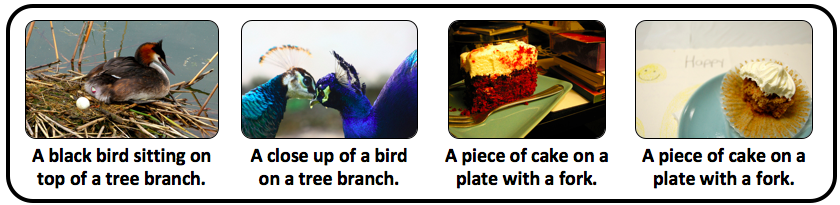}}
\label{fig:gencap}
\end{subfigure}\\[2px]
\begin{subfigure}[t]{\columnwidth}
\centering
\resizebox{\columnwidth}{0.8in}{\includegraphics{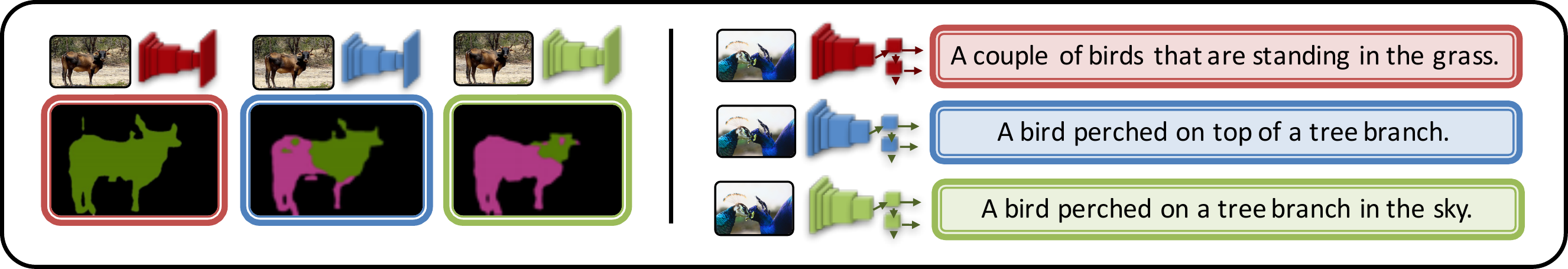}}
\end{subfigure}\\

\caption{Single-prediction based models often produce solutions with low \emph{expected loss} in the face of ambiguity; however, these solutions are often unrealistic or do not reflect the image content well (row 1). Instead, we train ensembles under a unified loss which allows each member to produce different outputs reflecting multi-modal beliefs (row 2). We evaluate our method on image classification, segmentation, and captioning tasks.}
\label{fig:teaser}
\vspace{-15pt}
\end{figure}

Diverse solution sets are especially useful for 
structured prediction problems with multiple reasonable interpretations, only
one of which is correct. Situations that often arise in practical systems include:
\begin{compactenum}[--]
\item \textbf{Implicit class confusion.} The label space of many classification
problems is often an arbitrary quantization of a continuous space.
For example, a vision system may be expected to classify between tables and desks, despite
many real-world objects arguably belonging to both classes.
By making multiple predictions, this implicit confusion can be viewed explicitly in system outputs. 
\item \textbf{Ambiguous evidence.} Often there is simply not enough information to make
a definitive prediction. For example, even a human expert may not be able to
identify a fine-grained class (e.g., particular breed of dog) given an occluded or distant view,
but they likely can produce a small set of reasonable guesses.
In such cases, the task of producing a diverse set of possibilities is more clearly defined than producing one correct answer.
\item \textbf{Bias towards the mode.} Many models have a tendency to exhibit 
mode-seeking behaviors as a way to reduce expected loss over a dataset (e.g., a conversation
model frequently producing `I don't know'). By making multiple predictions, a system can improve coverage of lower density areas
of the solution space, without sacrificing performance on the majority of examples.
\end{compactenum}
In other words, by optimizing for the oracle loss, a multiple-prediction learner can respond to
ambiguity much like a human does, by making multiple guesses that capture
multi-modal beliefs. 
In contrast, a single-prediction learner is forced to produce
a solution with low expected loss in the face of ambiguity. 
Figure \ref{fig:teaser} illustrates how this can 
produce solutions that are not useful in practice.
In semantic segmentation, for example, 
this problem often causes objects to be predicted 
as a mixture of multiple classes (like the horse-cow
shown in the figure).  In image captioning, minimizing expected loss
encourages generic sentences that are
`safe' with respect to expected error but
not very informative. For example, Figure \ref{fig:teaser} shows two pairs of images
each having different image content but very similar, generic captions --
the model knows it is safe to assume that birds are on branches
and that cakes are eaten with forks.

In this paper, we generalize the Multiple Choice Learning paradigm \cite{rivera_nips12,rivera_aistats14} to jointly learn ensembles of deep networks that \emph{minimize the oracle loss directly}. 
We are the first to adapt these ideas to deep networks and we present a novel training algorithm that avoids
costly retraining \cite{rivera_nips12} and learning difficulty \cite{Dey_2015_ICCV} of past methods.
Our primary technical contribution is the formulation of a stochastic block gradient descent optimization approach 
well-suited to minimizing the oracle loss in ensembles of deep networks, which we call \textbf{Stochastic Multiple Choice 
Learning (sMCL)}. Our formulation is applicable to any model trained with stochastic gradient descent, 
is agnostic to the form of the task dependent loss, is parameter-free, and is time efficient, training all ensemble members concurrently. 

We demonstrate the broad applicability and efficacy of sMCL for training diverse deep ensembles with \emph{interpretable emergent expertise} on a wide range of problem domains and network architectures, including Convolutional Neural Network (CNN) \cite{cifar-quick} ensembles for image classification \cite{cifar10}, Fully-Convolutional Network (FCN) \cite{long2015fully} ensembles for semantic segmentation \cite{pascal-voc-2011}, and combined CNN and Recurrent Neural Network (RNN) ensembles \cite{karpathy2015deep} for image captioning \cite{coco}. We provide detailed analysis of the training and output 
behaviors of the resulting ensembles, demonstrating how ensemble member specialization and expertise \emph{emerge} automatically when trained using sMCL. Our method outperforms existing baselines and produces sets of outputs with high oracle performance.

\section{Related Work}
\label{sec:related}

\xhdr{Ensemble Learning.} Much of the existing work on training ensembles focuses on diversity between member models as a means to improve performance by decreasing error correlation. This is often accomplished by resampling existing training data for each member model \cite{tumer1996error} or by producing artificial data that encourages new models to be decorrelated with the existing ensemble \cite{melville2005creating}. Other approaches train or combine ensemble members under a joint loss \cite{strehl2003cluster,liu1999ensemble}. More recently, work of Hinton et al.~\cite{hinton2014distilling} and Ahmed et al.~\cite{ahmed16experts} explores using `generalist' network performance statistics to inform the design of ensemble-of-expert architectures for classification. In contrast, sMCL \emph{discovers} specialization as a consequence of minimizing oracle loss. Importantly, most existing methods do not generalize to structured output labels, while sMCL seamlessly adapts, discovering different task-dependent specializations automatically.

\xhdr{Generating Multiple Solutions.} There is a large body of work on the topic of extracting multiple diverse solutions from a single model \cite{batra_eccv12,kirillov2015mbest,park2011n,kirillov2015mbesticcv,prasad_nips14}; however, these approaches are designed for probabilistic structured-output models and are not directly applicable to general deep architectures. 

Most related to our approach is the work of Guzman-Rivera et al.~\cite{rivera_nips12,rivera_aistats14} which explicitly minimizes oracle loss over the outputs of an ensemble, formalizing this setting as the Multiple Choice Learning (MCL) paradigm. They introduce a general alternating block coordinate descent training approach which requires retraining models multiple times. More recently, Dey et al.~\cite{Dey_2015_ICCV} reformulated this problem as a submodular optimization task in which ensemble members are learned sequentially in a boosting-like manner to maximize marginal gain in oracle performance. Both these methods require either costly retraining or sequential training, making them poorly suited to modern deep architectures that can take weeks to train. To address this serious shortcoming and to provide the first practical algorithm for training diverse deep ensembles, we introduce a stochastic gradient descent (SGD) based algorithm to train ensemble members concurrently.

\section{Multiple-Choice Learning as Stochastic Block Gradient Descent}
\label{method}
We consider the task of training an ensemble of differentiable learners that 
together produce a set of solutions with minimal loss with respect to an oracle 
that selects only the lowest-error prediction.

\xhdr{Notation.} We use $[n]$ to denote the set $\{1,2,\ldots,n\}$.
Given a training set of input-output pairs $D=\{(x_i,y_i)\mbox{ | }x_i
\in \mathcal{X}, y_i\in \mathcal{Y}\}$, our goal is to learn a function
$g: \mathcal{X} \rightarrow \mathcal{Y}^M$ which maps each input to $M$ outputs.
We fix the form of $g$ to be an ensemble of $M$ learners $f$ such that 
$g(x) = (f_1(x), \dots, f_M(x))$. For some task-dependent loss 
$\ell(y,\hat{y})$, which measures the error between true and predicted outputs 
$y$ and $\hat{y}$, we define the oracle loss of $g$ over the dataset $D$ as
$$\mathcal{L}_{O}(D) = \sum_{i=1}^n\min_{m \in [M]} \ell\left(y_i,f_m(x_i)\right).$$

\xhdr{Minimizing Oracle Loss with Multiple Choice Learning.} In order to directly minimize the oracle loss for an 
ensemble of learners, Guzman-Rivera et al.~\cite{rivera_nips12} present an objective 
which forms a (potentially tight) upper-bound. This objective replaces the $\min$ in 
the oracle loss with indicator variables $(p_{i,m})_{m=1}^M$ where $p_{i,m}$ is 1 if predictor $m$ has 
the lowest error on example $i$, 
\begin{eqnarray}
\label{eq:oloss} \argmin_{f_m,p_{m,i}} && \sum_{i=1}^n\sum_{m=1}^M p_{i,m}\mbox{ }\ell\left(y_i,f_m(x_i)\right) \\
 \st&& \quad \sum^M p_{i,m} = 1, \;  \quad p_{i,m} \in \{ 0, 1 \}. \nonumber
\end{eqnarray}
The resulting minimization is a constrained joint optimization over ensemble parameters and data-point assignments.
The authors propose an alternating block algorithm, shown in Algorithm 1, to approximately 
minimize this objective. Similar to K-Means or `hard-EM,' this approach alternates between 
assigning examples to their min-loss predictors and training models to convergence
on the partition of examples assigned to them. Note that this approach is not feasible with training deep networks,
since modern architectures~\cite{he2015deep} can take weeks or months to train a
single model once.

\begin{figure}
\includegraphics[width=\columnwidth]{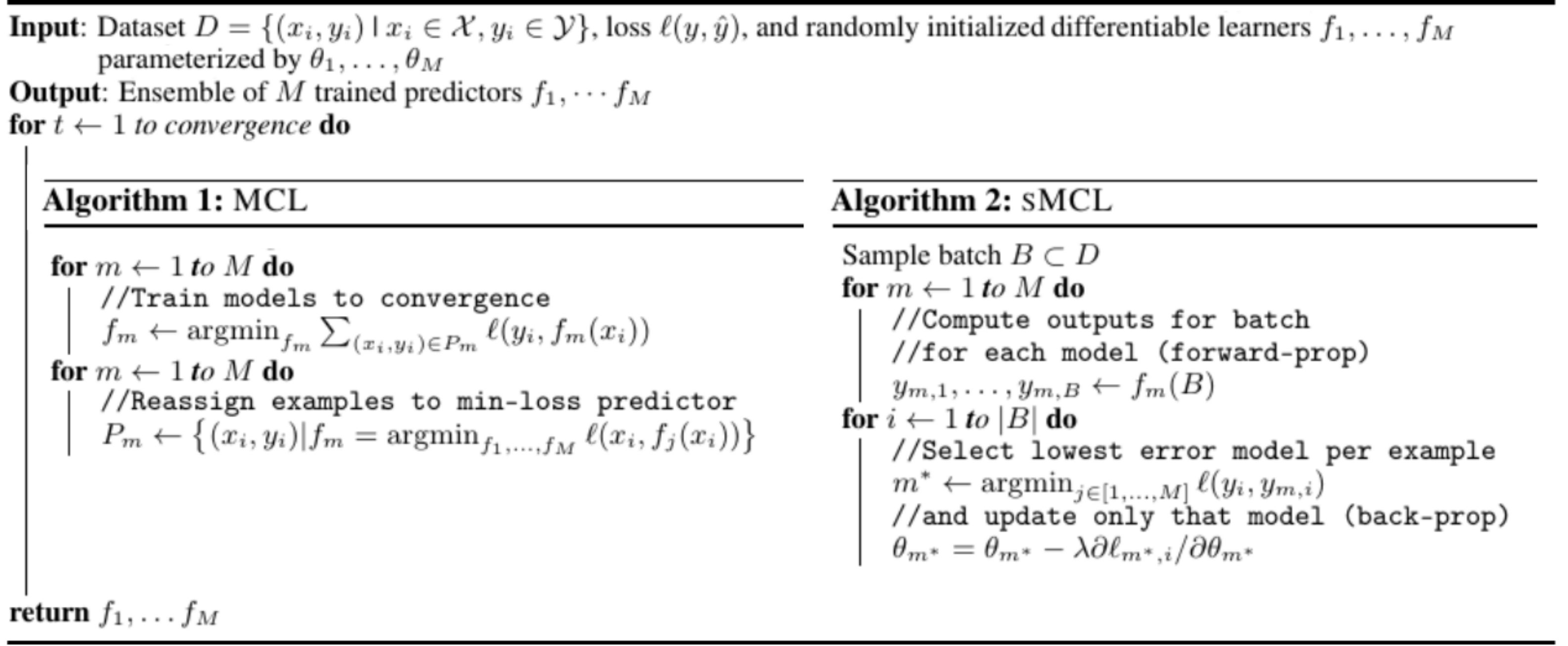}\\
\caption{The MCL approach of \cite{rivera_nips12} (Alg. 1) requires costly retraining while our sMCL method (Alg. 2) works within standard SGD solvers, training all ensemble members under a joint loss.}
\end{figure}

\xhdr{Stochastic Multiple Choice Learning.} To overcome this shortcoming, we propose a stochastic algorithm for differentiable learners which 
interleaves the assignment step with batch updates in stochastic gradient descent. Consider the partial 
derivative of the objective in Eq.~\ref{eq:oloss} with respect to the output of the $m^{th}$ individual learner on example $x_i$,
\begin{equation}
\frac{\partial \mathcal{L}_O}{\partial f_m(x_i)} = p_{i,m}\mbox{ }\frac{\partial\ell(y_i, f_m(x_i))}{\partial f_m(x_i)}. \label{eq:deriv}
\end{equation}
Notice that if $f_m$ is the minimum error predictor for example $x_i$, then $p_{i,m}=1$, and the gradient term 
is the same as if training a single model; otherwise, the gradient is zero. This behavior lends itself
to a straightforward optimization strategy for learners trained by SGD based solvers. 
For each batch, we pass the examples through the learners, calculating losses from each ensemble member for each example. 
During the backward pass, the gradient of the loss for each example is backpropagated only to the lowest error predictor on that example (with ties broken arbitrarily). 

This approach, which we call Stochastic
Multiple Choice Learning (sMCL), is shown in Algorithm 2. sMCL is generalizable to any learner 
trained by stochastic gradient descent and is thus applicable to an extensive range of modern deep 
networks. Unlike the iterative training schedule of MCL, sMCL ensembles need only be trained to convergence 
once in parallel. sMCL is also agnostic to the exact form of loss function $\ell$ such that it can be applied 
without additional effort on a variety of problems.

\section{Experiments}
\label{experiments}

\begin{figure}[t]
\captionsetup[subfigure]{aboveskip=5pt}
\begin{subfigure}[b]{0.3\columnwidth}
\centering
\resizebox{0.8\columnwidth}{!}{\includegraphics[clip=true, trim=0px 8px 0px 0px]{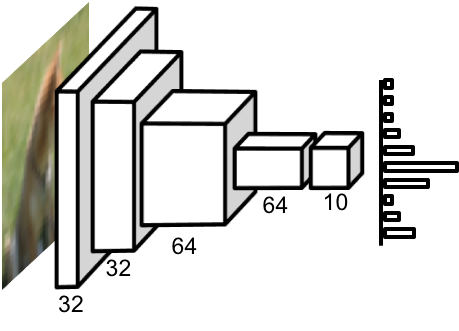}}
\caption{Convolutional classification model of \cite{cifar-quick} for CIFAR10~\cite{cifar10}}
\label{fig:cifar10}
\end{subfigure}
\hfill
\begin{subfigure}[b]{0.30\columnwidth}
\centering
\resizebox{\columnwidth}{!}{\includegraphics{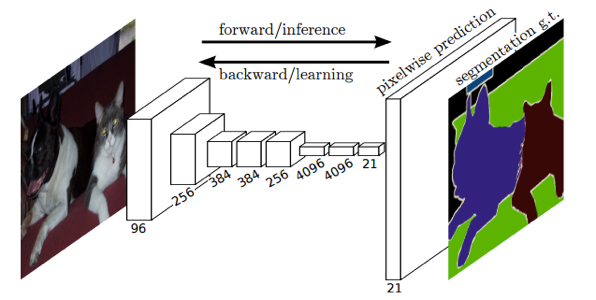}}
\caption{Fully-convolutional segmentation model of Long et al.~\cite{long2015fully}}
\label{fig:segment}
\end{subfigure}
\hfill
\begin{subfigure}[b]{0.30\columnwidth}
\centering
\resizebox{0.9\columnwidth}{!}{\includegraphics{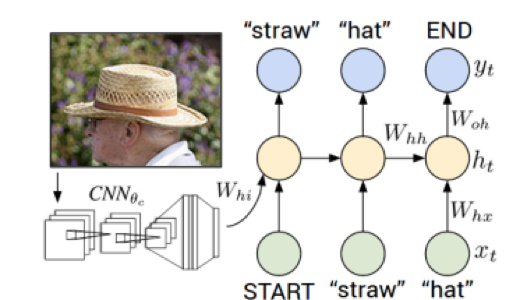}}
\caption{CNN+RNN based captioning model of Karpathy et al.~\cite{karpathy2015deep}}
\label{fig:caption}
\end{subfigure}
\vspace{2pt}
\caption{We experiment with three problem domains using the various architectures shown above.}
\label{fig:models}
\end{figure}

In this section, we present results for sMCL ensembles trained for the tasks and deep architectures
shown in Figure \ref{fig:models}. These include CNN ensembles for image classification, FCN ensembles for semantic segmentation, and a CNN+RNN ensembles for image caption generation. 

\xhdr{Baselines.} Many existing general techniques for inducing diversity are not directly applicable to deep networks.
We compare our proposed method against:
\begin{compactenum}[-]
\item \textbf{Classical} ensembles in which each model is trained under an independent loss with differing random initializations. We will refer to these as \textbf{Indp.} ensembles in figures.

\item \textbf{MCL \cite{rivera_nips12}} that alternates between training models to convergence on assigned examples and allocating
examples to their lowest error model. We repeat this process for 5 meta-iterations and initialize ensembles with (different) random weights. 
We find MCL performs similarly to sMCL on small classification tasks; however, MCL performance drops substantially on segmentation and captioning tasks. 
Unlike sMCL which can effectively reassign an example once per epoch, MCL only does this after convergence, limiting its capacity to specialize
compared to sMCL. We also note that sMCL is 5x faster than MCL, where the factor 5 is the result of choosing 5 meta-iterations (other applications may require more, further increasing the gap.)

\item \textbf{Dey et al.~\cite{Dey_2015_ICCV}} train models sequentially in a boosting-like fashion, each time reweighting examples to maximize marginal increase of the
evaluation metric. We find these models saturate quickly as the ensemble size grows. As performance increases, the marginal gain and 
therefore the weights approach zero. With low weights, the average gradient backpropagated for stochastic learners drops substantially,
reducing the rate and effectiveness of learning without careful tuning. To compute weights, \cite{Dey_2015_ICCV} requires an error measure
bounded above by 1: accuracy (for classification) and IoU (for segmentation) satisfy this; the CIDEr-D score~\cite{Vedantam_2015_CVPR} divided by 10 guarantees this for captioning.
\end{compactenum}

\xhdr{Oracle Evaluation.} We present results as oracle versions of the task-dependent
performance metrics. These oracle metrics report the highest score over all outputs for a given input. For 
example, in classification tasks, oracle accuracy is exactly the top-$k$ criteria of ImageNet~\cite{ilsvrc12}, \ie whether at least one of the outputs is the correct label. Likewise, the oracle intersection over union (IoU) is the highest IoU between the ground
truth segmentation and any one of the outputs. Oracle metrics allow the evaluation of multiple-prediction systems
separately from downstream re-ranking or selection systems, and have been extensively used in previous work~\cite{rivera_nips12,rivera_aistats14,batra_eccv12,kirillov2015mbest,park2011n,kirillov2015mbesticcv,prasad_nips14,Dey_2015_ICCV}.

Our experiments convincingly demonstrate the broad applicability and efficacy of sMCL for training diverse deep ensembles. In all three experiments, sMCL significantly outperforms classical ensembles, Dey et al.~\cite{Dey_2015_ICCV} (typical improvements of 6-10\%), and MCL (while providing a 5x speedup over MCL). Our analysis shows that the exact same algorithm (sMCL) leads to the automatic emergence of different interpretable notions of specializations among ensemble members.

\subsection{Image Classification}
\label{classification}

\xhdr{Model.} We begin our experiments with sMCL on the CIFAR10~\cite{cifar10} dataset using the small convolutional neural network \emph{``CIFAR10-Quick''} provided with the Caffe deep learning framework~\cite{caffe}. CIFAR10 is a ten way classification task with small 32$\times$32 images. For these experiments, the reference model is trained using a batch size of 350 for 5,000 iterations with a momentum of 0.9, weight decay of 0.004, and an initial learning rate of 0.001 which drops to 0.0001 after 4000 iterations. 

\begin{figure}[t]
\captionsetup[subfigure]{aboveskip=1pt, belowskip=1pt}
\centering
\begin{subfigure}[t]{0.49\columnwidth}
\begin{tikzpicture}
\begin{axis}[
    xlabel={Ensemble Size $M$},
    ylabel={Oracle Accuracy},
    xtick={1,2,3,4,5,6},
    xmin=0.5, xmax=6.5,
    ymin=76, ymax=98,
    legend pos=south east,
    legend style={font=\fontsize{5}{0}\selectfont},
    legend columns=2
    ymajorgrids=true,
    grid style=dashed,
    ylabel style={yshift=-1.2em},
    xlabel style={yshift=0.5em},
    width=2.8in,
    height=1.5in,
]

\addplot[
    color=blue,
    mark=square,
    ]
    coordinates {(1,77.11)(2,85.47)(3,88.65)(4,93.1)(5,94.29)(6,96.2)};
    \addlegendentry{sMCL}

\addplot[
    color=green,
    mark=triangle,
    ]
    coordinates {(1,77.22) (2,84.69) (3,88.44) (4,92.09)(5,94.64)(6,95.53)};
    \addlegendentry{MCL}

\addplot[
    color=cyan,
    mark=star,
    ]
    coordinates {(1,77.11)(2, 83.3)(3,86.04)(4,87.35)(5,88.25)(6,88.84)};
    \addlegendentry{Dey \cite{Dey_2015_ICCV}}

\addplot[
    color=red,
    mark=o,
    ]
    coordinates {(1,77.11)(2, 83.03)(3,86.58)(4,88.51)(5,90.09)(6,92.33)};
    \addlegendentry{Indp.}

\end{axis}
\end{tikzpicture}

\caption{Effect of Ensemble Size}
\label{fig:cifaracc}
\end{subfigure}
\hfill
\begin{subfigure}[t]{0.49\columnwidth}

\begin{tikzpicture}
\begin{axis}[
    xlabel={Iterations},
    ylabel={Oracle Loss},
    xtick={0,2500,5000},
    xmin=0, xmax=5000,
    ymin=0, ymax=5,
    legend pos=north east,
    legend style={font=\fontsize{5}{0}\selectfont},
    ymajorgrids=false,
    ylabel style={yshift=-1.6em},
    xlabel style={yshift=0.5em},
    width=2.8in,
    height=1.5in,
]
 
\addplot[
    color=blue,
    mark=none,
    ]
    coordinates {(0,9.20185)(50,2.19238)(100,1.7351)(150,1.715)(200,1.86816)(250,1.5623)(300,1.96897)(350,1.49581)(400,1.3912)(450,1.37233)(500,1.13929)(550,1.23701)(600,1.05816)(650,1.07033)(700,1.13862)(750,0.893673)(800,0.93573)(850,1.11479)(900,0.88104)(950,1.41076)(1000,0.775165)(1050,1.1843)(1100,1.05342)(1150,0.79895)(1200,0.854804)(1250,0.920951)(1300,0.878591)(1350,0.860291)(1400,0.802927)(1450,0.83192)(1500,0.596449)(1550,0.750438)(1600,0.727023)(1650,0.635202)(1700,0.633851)(1750,0.702509)(1800,0.652514)(1850,0.550898)(1900,0.499146)(1950,0.945256)(2000,0.567293)(2050,0.753184)(2100,0.794419)(2150,0.505931)(2200,0.601633)(2250,0.668324)(2300,0.672161)(2350,0.537823)(2400,0.57232)(2450,0.579246)(2500,0.408685)(2550,0.492726)(2600,0.51744)(2650,0.438338)(2700,0.352334)(2750,0.487222)(2800,0.346737)(2850,0.362617)(2900,0.526559)(2950,0.70175)(3000,0.432615)(3050,0.476397)(3100,0.409441)(3150,0.383566)(3200,0.391982)(3250,0.420836)(3300,0.405381)(3350,0.47958)(3400,0.326026)(3450,0.33518)(3500,0.336518)(3550,0.326704)(3600,0.30773)(3650,0.258918)(3700,0.264953)(3750,0.236177)(3800,0.27998)(3850,0.275607)(3900,0.182939)(3950,0.381865)(4000,0.316037)(4050,0.405491)(4100,0.310565)(4150,0.131862)(4200,0.167452)(4250,0.219727)(4300,0.206412)(4350,0.158768)(4400,0.12485)(4450,0.173497)(4500,0.128942)(4550,0.0999057)(4600,0.142523)(4650,0.111296)(4700,0.0985805)(4750,0.134803)(4800,0.0804806)(4850,0.113992)(4900,0.0905068)(4950,0.191675)};
    \addlegendentry{sMCL}

\addplot[
    color=red,
    mark=none,
    ]
    coordinates {(0,9.19612)(50,6.08642)(100,4.91222)(150,4.46244)(200,4.328346)(250,3.945044)(300,3.891214)(350,3.415885)(400,3.009441)(450,3.107462)(500,2.602751)(550,2.705355)(600,2.623476)(650,2.111689)(700,2.329667)(750,2.507018)(800,2.087141)(850,2.037013)(900,2.061022)(950,2.321032)(1000,1.977846)(1050,2.19836)(1100,2.019189)(1150,1.750736)(1200,1.780648)(1250,1.854328)(1300,1.861927)(1350,1.669426)(1400,1.488606)(1450,1.620703)(1500,1.484043)(1550,1.623298)(1600,1.472655)(1650,1.413518)(1700,1.397344)(1750,1.575614)(1800,1.383037)(1850,1.300881)(1900,1.37255)(1950,1.503069)(2000,1.379974)(2050,1.360495)(2100,1.32979)(2150,1.055984)(2200,1.292316)(2250,1.363949)(2300,1.355169)(2350,1.138793)(2400,1.010355)(2450,1.016485)(2500,1.159761)(2550,1.067476)(2600,1.099781)(2650,0.994493)(2700,0.99896)(2750,1.086394)(2800,0.935128)(2850,0.931382)(2900,0.900883)(2950,1.087226)(3000,1.016324)(3050,1.006244)(3100,1.051131)(3150,0.68932)(3200,0.961898)(3250,1.024148)(3300,1.091762)(3350,0.846709)(3400,0.757573)(3450,0.812978)(3500,0.821286)(3550,0.791807)(3600,0.798889)(3650,0.70087)(3700,0.717013)(3750,0.737702)(3800,0.698503)(3850,0.719885)(3900,0.631065)(3950,0.77201)(4000,0.73434)(4050,0.844298)(4100,0.725144)(4150,0.5153269)(4200,0.722193)(4250,0.84066)(4300,0.786271)(4350,0.700132)(4400,0.50279)(4450,0.626166)(4500,0.6564515)(4550,0.604838)(4600,0.688043)(4650,0.5565319)(4700,0.5664328)(4750,0.639985)(4800,0.5431725)(4850,0.536615)(4900,0.54959)(4950,0.649155)};
    \addlegendentry{Indp.}
 
\end{axis}
\end{tikzpicture}
\caption{Oracle Loss During Training $(M=4)$}
\label{fig:cifarloss}
\end{subfigure}
\\
\hspace{-10pt}
\begin{subfigure}{0.26\columnwidth}

\includegraphics[clip=true, trim=0px 10px 35px 20px, scale=0.22]{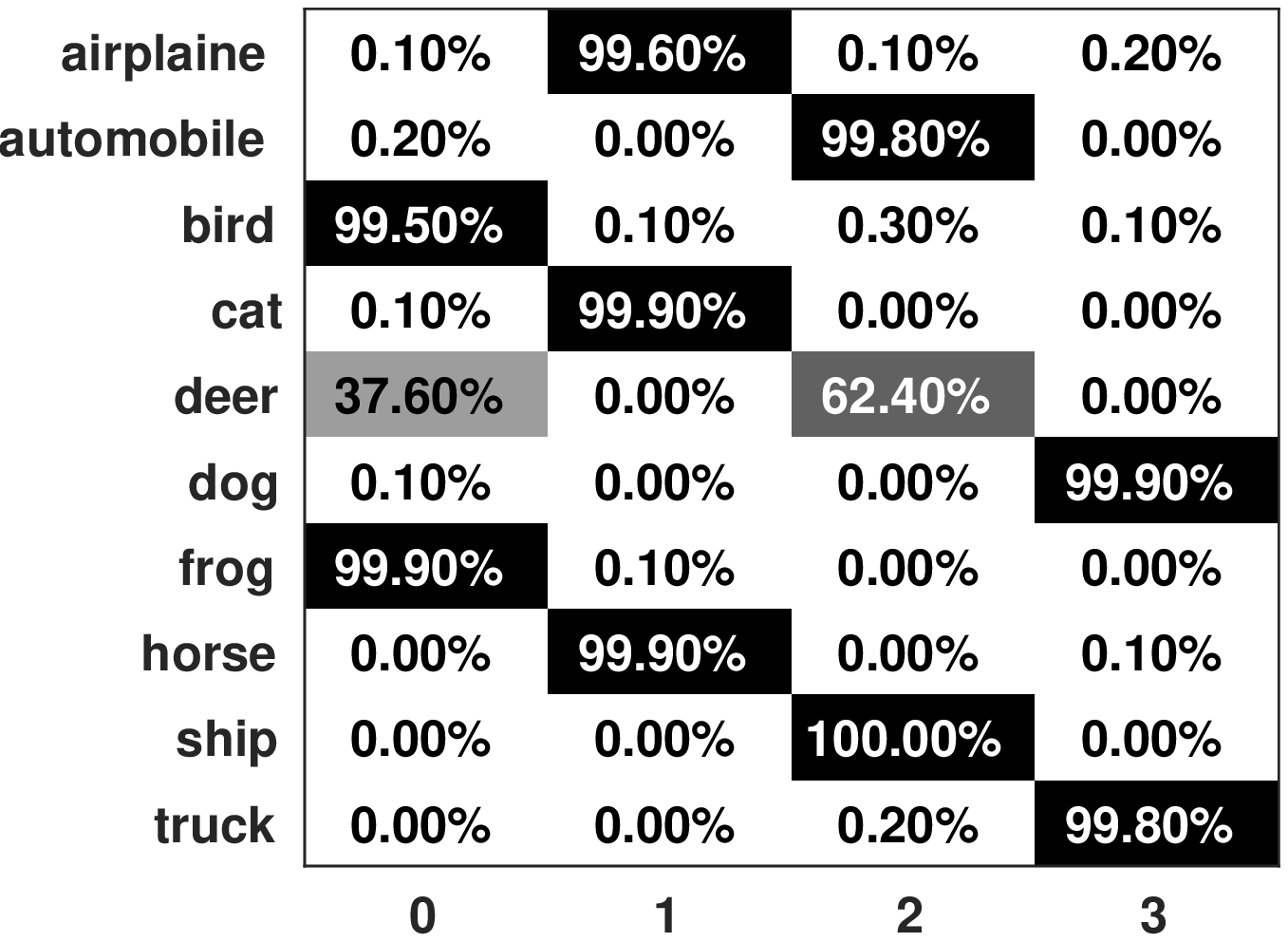}\null\hfill
\caption{k=1}
\label{fig:k1}
\end{subfigure}
\begin{subfigure}{0.23\columnwidth}
\centering
\includegraphics[clip=true, trim=40px 10px 35px 20px, scale=0.22]{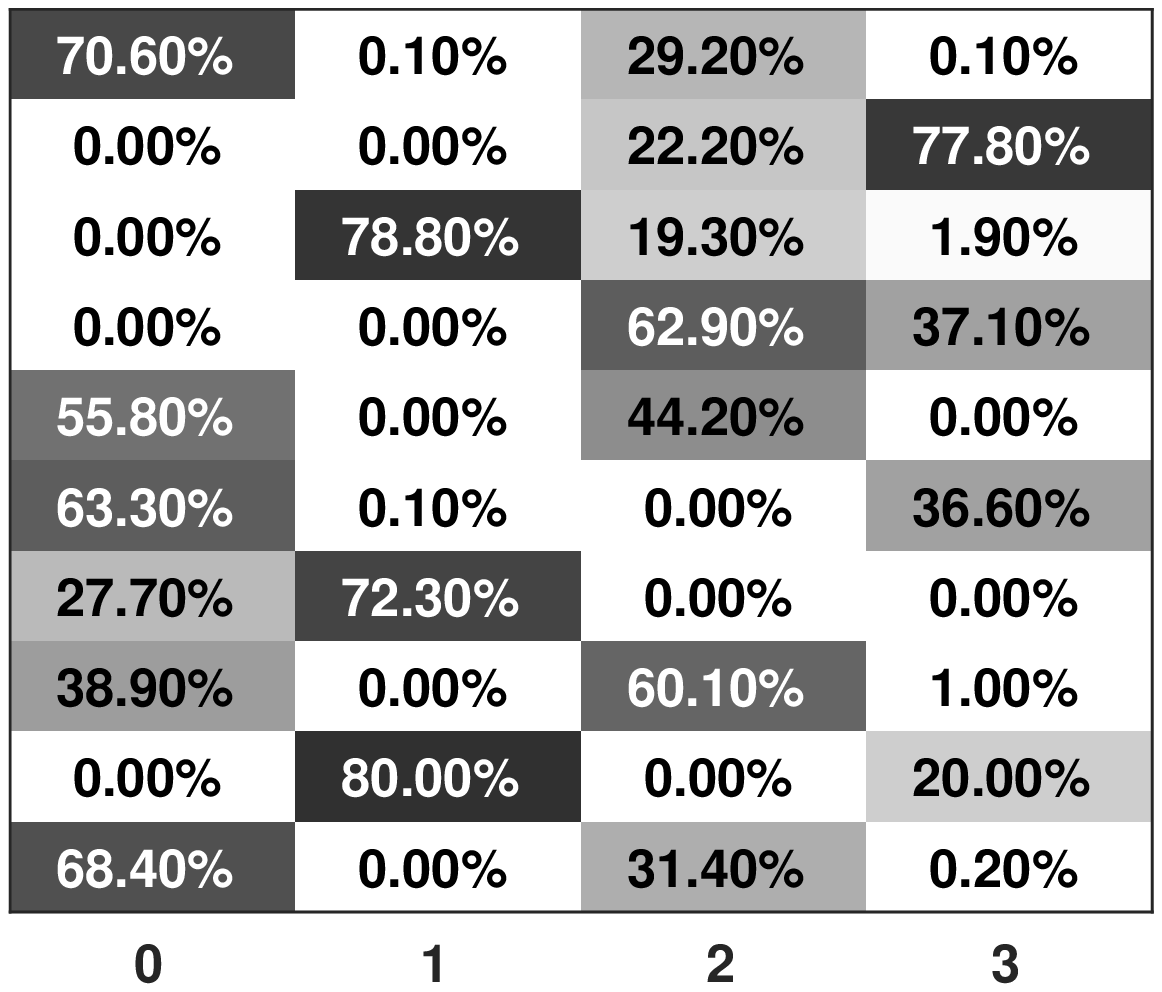}
\caption{k=2}
\label{fig:k2}
\end{subfigure}
\begin{subfigure}{0.23\columnwidth}
\centering
\includegraphics[clip=true, trim=40px 10px 35px 20px, scale=0.22]{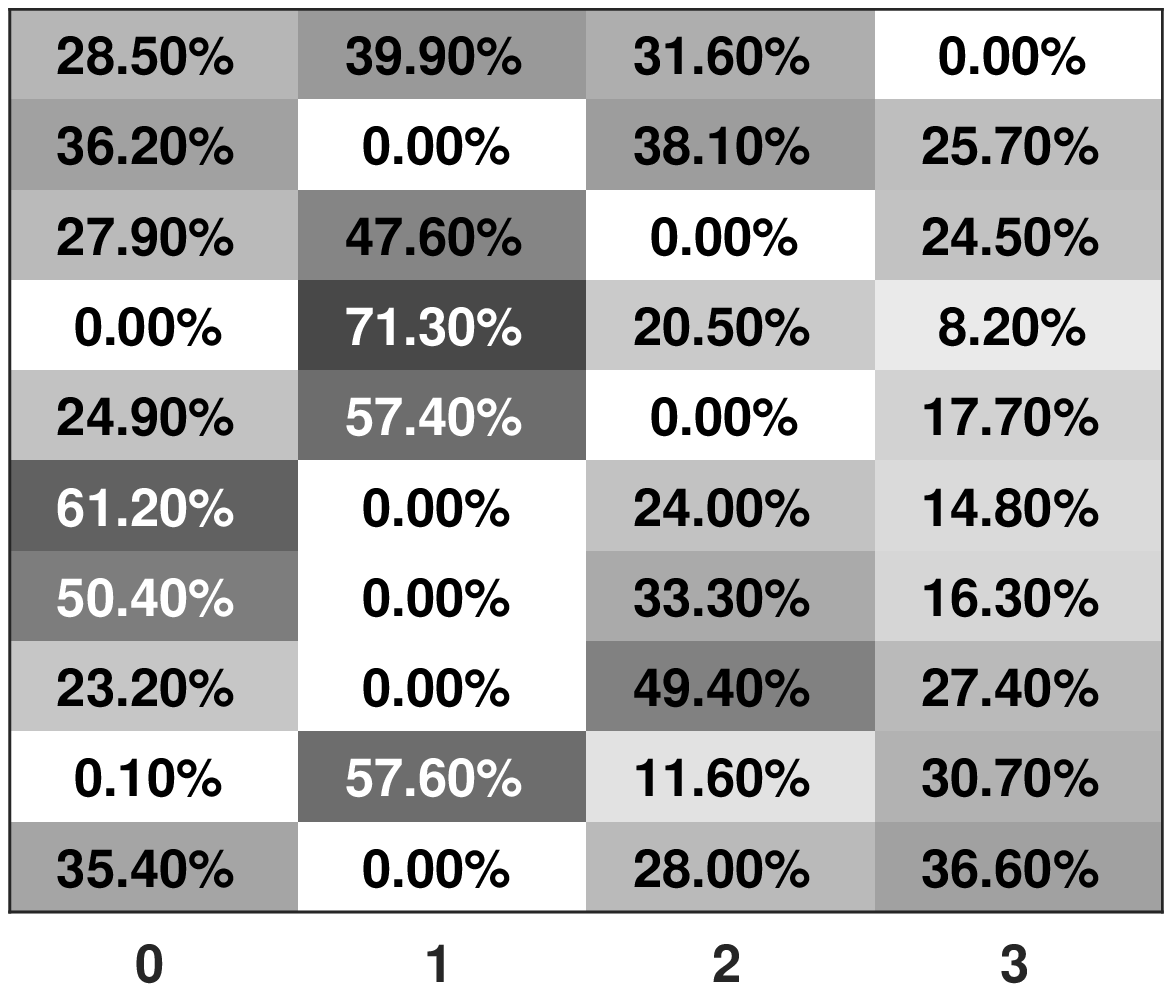}\null\hfill
\caption{k=3}
\label{fig:k3}
\end{subfigure}
\begin{subfigure}{0.23\columnwidth}
\centering
\includegraphics[clip=true, trim=40px 10px 35px 20px, scale=0.22]{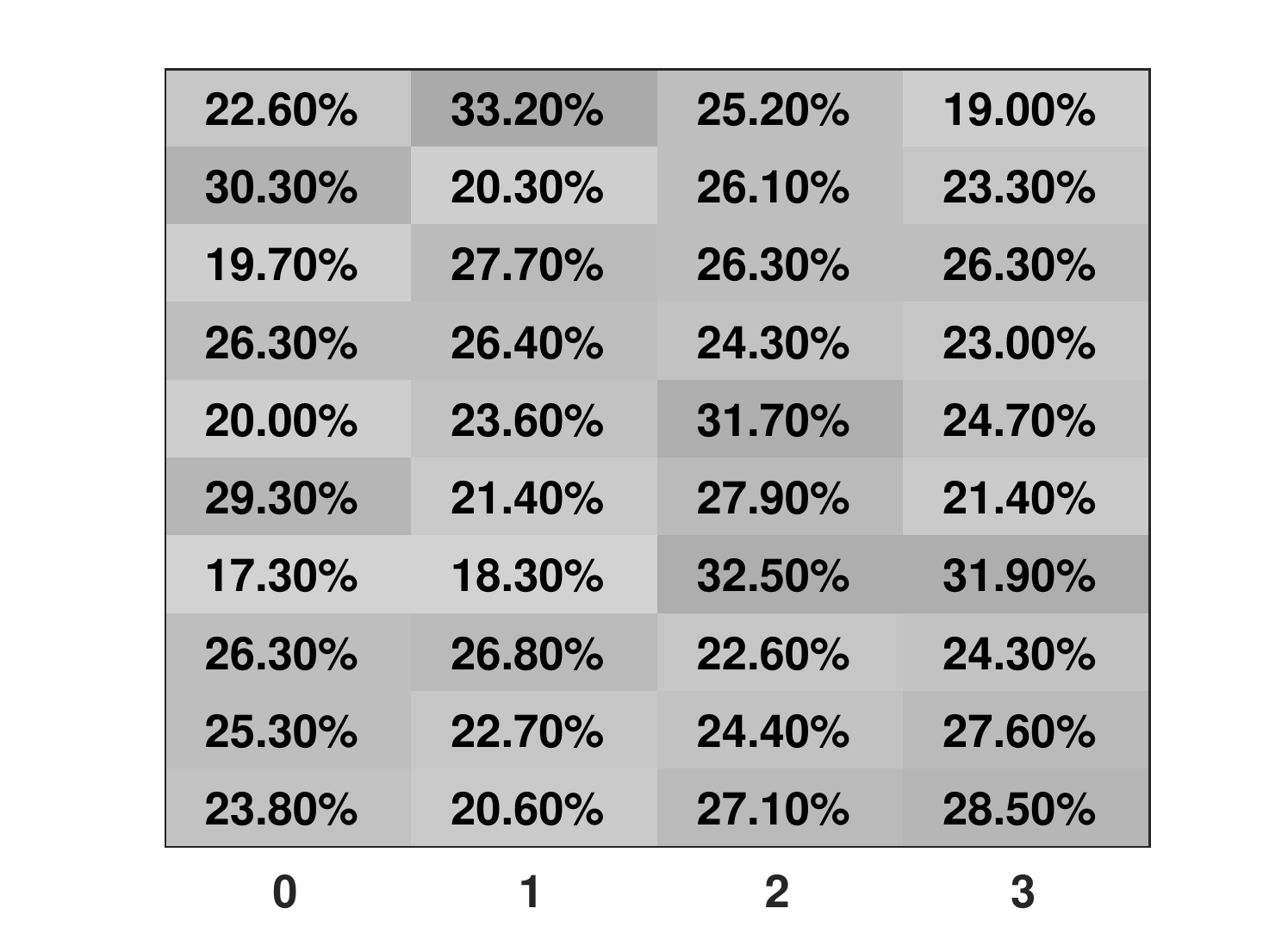}
\caption{k=M=4}
\label{fig:k4}
\end{subfigure}

\caption{sMCL trained ensembles produce higher oracle accuracies than baselines (a) by directly optimizing the oracle loss (b). By varying the number of predictors $k$ each example can be assigned to, we can interpolate between sMCL and standard ensembles, and (c-f) show the percentage of test examples of each class assigned to each ensemble member by the oracle for various $k$. These divisions are not preselected and show how specialization is an emergent property of sMCL training.}
\label{fig:cifarres}
\end{figure}

\xhdr{Results.} Oracle accuracy for sMCL and baseline ensembles of size 1 to 6 are shown in Figure \ref{fig:cifaracc}. The sMCL trained ensembles result in higher oracle accuracy than the baseline methods, and are comparable to MCL while being 5x faster. The method of Dey et al.~\cite{Dey_2015_ICCV} performs worse than independent ensembles as ensemble size grows. Figure \ref{fig:cifarloss} shows the oracle loss during training for sMCL and regular ensembles. The sMCL trained models optimize for the oracle cross-entropy loss directly, not only arriving at lower loss solutions but also reducing error more quickly. 

\xhdr{Interpretable Expertise: sMCL Induces Label-Space Clustering.} Figure \ref{fig:k1} shows the class-wise distribution of the assignment of test datapoints to the oracle or `winning' predictor for an $M=4$ sMCL ensemble. The level of class division is \emph{striking} -- most predictors \emph{become specialists for certain classes}.
Note that these divisions emerge from training under the oracle loss and are not hand-designed or pre-initialized in any way. In contrast, Figure \ref{fig:k4} show that the oracle assignments for a standard ensemble are nearly uniform. To explore the space between these two extremes, we loosen the constraints of Eq. \ref{eq:oloss} such that the lowest $k$ error predictors are penalized. By varying $k$ between 1 and the number of ensemble members $M$, the models transition from minimizing oracle loss at $k=1$ to a traditional ensemble at $k=M$. Figures \ref{fig:k2} and \ref{fig:k3} show these results. We find a direct correlation between the degree of specialization and oracle accuracy, with $k=1$ netting highest oracle accuracy.


\subsection{Semantic Segmentation}
\label{segmentation}

We now present our results for the semantic segmentation task on the Pascal VOC dataset \cite{pascal-voc-2011}. 

\xhdr{Model.} We use the fully convolutional network (FCN) architecture presented by Long et al.~\cite{long2015fully} as our base model. Like \cite{long2015fully}, we train on the Pascal VOC 2011 training set augmented with extra segmentations provided in \cite{hariharan_iccv11} and we test on a subset of the VOC 2011 validation set. We initialize our sMCL models from a standard ensemble trained for 50 epochs at a learning rate of $10^{-3}$. The sMCL ensemble is then fine-tuned for another 15 epochs at a reduced learning rate of $10^{-5}$.

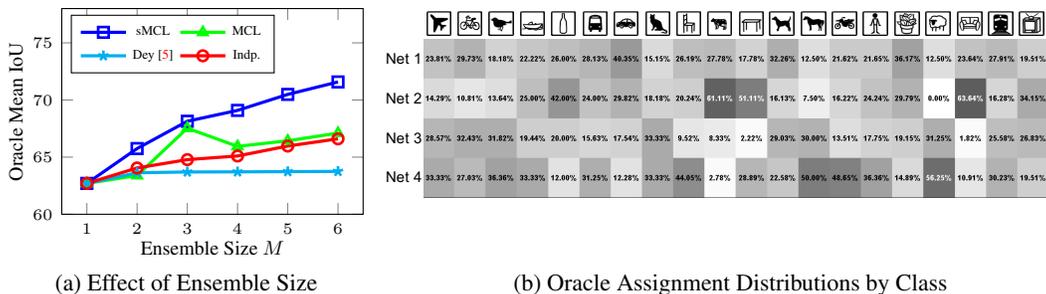
\begin{figure}[b]
\centering
\begin{subfigure}[t]{0.35\columnwidth}
\begin{tikzpicture}
\begin{axis}[
    xlabel={Ensemble Size $M$},
    ylabel={Oracle Mean IoU},
    xtick={1,2,3,4,5,6},
    xmin=0.5, xmax=6.5,
    ymin=60, ymax=78,
    legend pos=north west,
    legend style={font=\fontsize{5}{0}\selectfont},
    legend columns=2
    ymajorgrids=true,
    grid style=dashed,
    ylabel style={yshift=-1.2em},
    xlabel style={yshift=0.5em},
    width=2.2in,
    height=1.7in,
]
 
\addplot[
    color=blue,
    mark=square,
    ]
    coordinates {(1,62.7)(2,65.75)(3,68.14)(4,69.09)(5,70.49)(6,71.58)};
    \addlegendentry{sMCL}
\addplot[
    color=green,
    mark=triangle,
    ]
    coordinates {(1,62.7) (2,63.39) (3,67.53) (4,65.94)(5,66.43)(6,67.10)};
    \addlegendentry{MCL}
\addplot[
    color=cyan,
    mark=star,
    ]
    coordinates {(1,62.7)(2, 63.63)(3,63.70)(4,63.71) (5,63.73)(6,63.75)};
    \addlegendentry{Dey \cite{Dey_2015_ICCV}}
\addplot[
    color=red,
    mark=o,
    ]
    coordinates {(1,62.7)(2, 64.05)(3,64.78)(4,65.10)(5,65.97)(6,66.60)};
    \addlegendentry{Indp.}
\end{axis}
\end{tikzpicture}
\null\hfill
\caption{Effect of Ensemble Size}
\label{fig:pascaliou}
\end{subfigure}
\hspace{10px}
\begin{subfigure}[t]{0.61\columnwidth}
\hspace{-13pt}\resizebox{1.05\columnwidth}{!}{\includegraphics[trim=0 -58px 0 0]{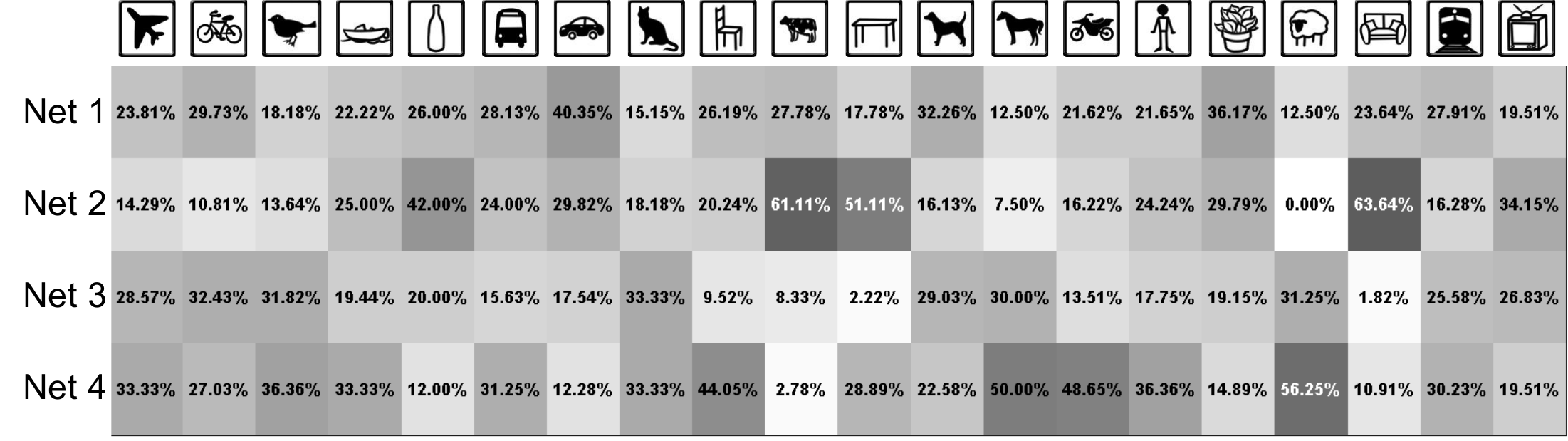}}
\caption{Oracle Assignment Distributions by Class}
\label{fig:pascaldiv}
\end{subfigure}
\caption{a) sMCL trained ensembles consistently result in improved oracle mean IoU over baselines on PASCAL VOC 2011. b) Distribution of examples from
each category assigned by the oracle for an sMCL ensemble.}
\end{figure}

\xhdr{Results.} Figure \ref{fig:pascaliou} shows oracle accuracy (class-averaged IoU) for all methods with ensemble sizes ranging from 1 to 6. Again, sMCL significantly outperforms all baselines (\textasciitilde$7\%$ relative improvement over classical ensembles). In this more complex setting, we see the method of Dey et al.~\cite{Dey_2015_ICCV} saturates more quickly -- resulting in performance worse than classical ensembles as ensemble size grows. Though we expect MCL to achieve similar results as sMCL, retraining the MCL ensembles a sufficient number of times proved infeasible so results after five meta-iterations are shown.

\xhdr{Interpretable Expertise: sMCL as Segmentation Specialists.} In Figure \ref{fig:pascaldiv}, we analyze the class distribution of the predictions using an sMCL ensemble with $4$ members. For each test sample, the oracle picks the prediction which corresponds to the ensemble member with the highest accuracy for that sample. We find the specialization with respect to classes is much less evident than in the classification experiments. As segmentation presents challenges other than simply selecting the correct class, specialization can occur in terms of shape and frequency of predicted segments in addition to class divisions; however, we do still see some class biases -- network 2 captures cows, tables, and sofas well and network 4 has become an expert on sheep and horses.

Figure \ref{fig:segqual} shows qualitative results from a four member sMCL ensemble. We can clearly observe the diversity in the segmentations predicted by different members. In the first row, we see the majority of the ensemble members produce dining tables of various completeness in response to the visual uncertainty caused by the clutter. Networks 2 and 3 capture this ambiguity well, producing segmentations with the dining table completely present or absent. Row 2 demonstrates the capacity of sMCL ensembles to provide multiple high quality solutions. The models are confused whether the animal is a horse or a cow -- models 1 and 3 produce typical `safe' responses while models 2 and 4 attempt to give cohesive responses. Finally, row 3 shows how the models can learn biases about the frequency of segments with model 3 presenting only the sheep.

\newcolumntype{C}[1]{>{\centering\let\newline\\\arraybackslash\hspace{0pt}}m{#1}}
\newcommand{\res}[1]{\resizebox{1.2in}{0.8in}{#1}}
\fboxsep=0.25mm
\fboxrule=0.7mm
\begin{figure}[t!]
\centering
\resizebox{0.9\columnwidth}{!}{
\setlength{\tabcolsep}{0.5pt}
\begin{tabular}{C{1.2in}@{\hskip 0.25in} C{1.2in}@{\hskip 0.15in} C{1.2in} C{1.2in} C{1.2in} C{1.2in}}
  & Independent\newline Ensemble Oracle &\multicolumn{4}{c}{sMCL Ensemble Predictions} \\ \cline{3-6}
\res{\includegraphics[width=1.1in,height=0.8in]{{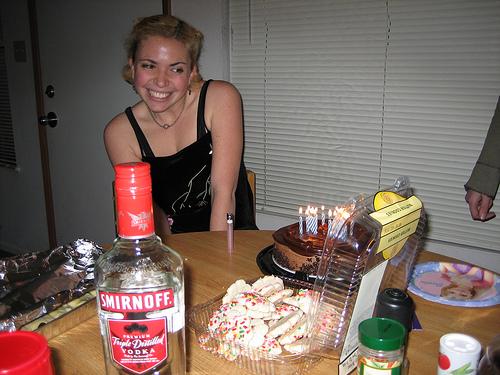}}}&
\res{\includegraphics[width=1.1in, height=0.8in, clip=true, trim= 80px 60px 70px 75px]{{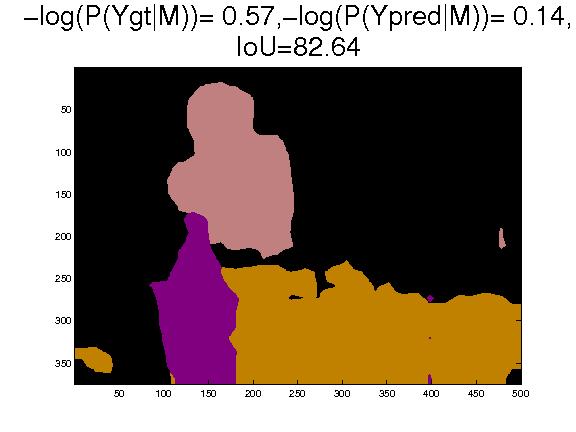}}}&
\res{\includegraphics[width=1.1in, height=0.8in, clip=true, trim= 72px 45px 40px 65px]{{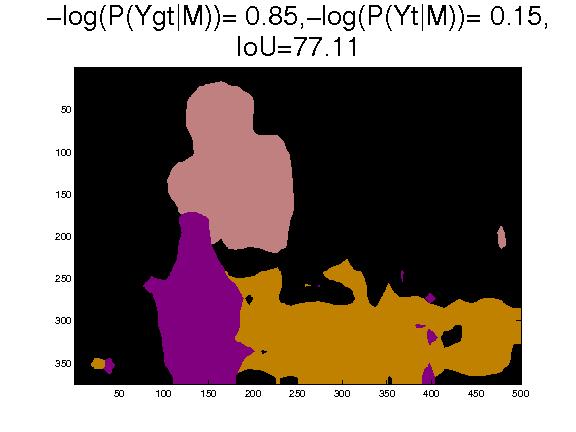}}}&
\raisebox{2pt}{\res{\fcolorbox{red}{white}{\includegraphics[width=1.1in, height=0.8in, clip=true, trim= 80px 60px 70px 75px]{{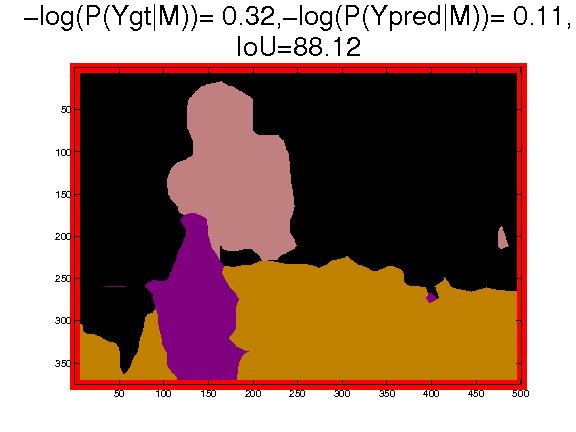}}}}}&
\res{\includegraphics[width=1.1in, height=0.8in, clip=true, trim= 72px 45px 40px 65px]{{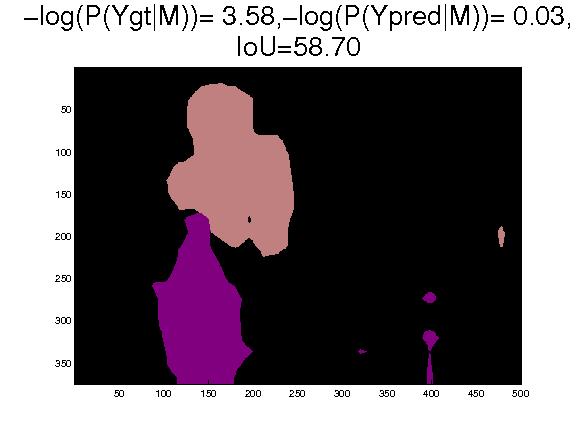}}}&
\res{\includegraphics[width=1.1in, height=0.8in, clip=true, trim= 72px 45px 40px 65px]{{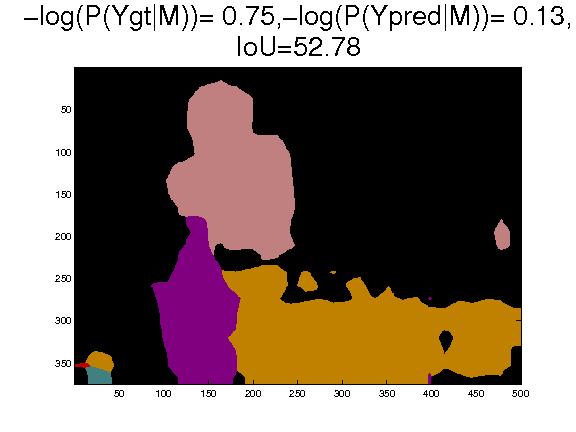}}}\\[-3px]
&  {\scriptsize IoU 82.64} & {\scriptsize IoU 77.11} & {\textbf{\scriptsize IoU 88.12}} & {\scriptsize IoU 58.70} & {\scriptsize IoU 52.78} \\[-7px]

\res{\includegraphics[width=1.1in,height=0.8in]{{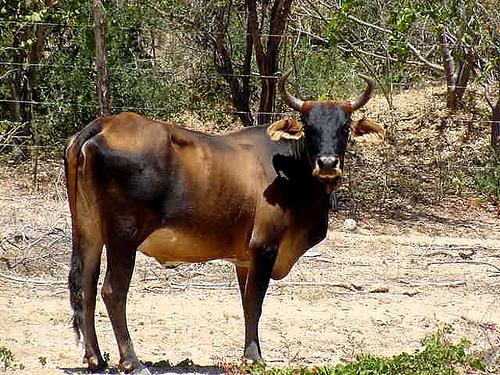}}}&
\res{\includegraphics[width=1.1in, height=0.8in, clip=true, trim= 80px 60px 70px 75px]{{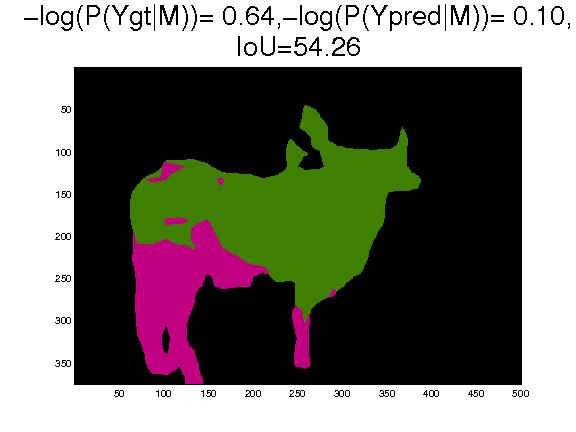}}}&
\res{\includegraphics[width=1.1in, height=0.8in,clip=true, trim= 72px 45px 40px 65px]{{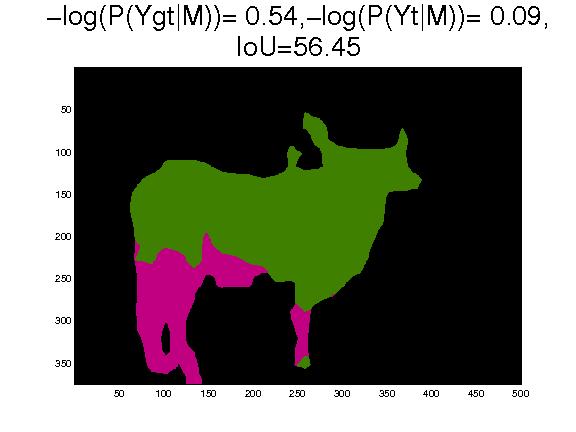}}}&
\raisebox{2pt}{\res{\fcolorbox{red}{white}{\includegraphics[width=1.1in, height=0.8in, clip=true, trim= 80px 60px 70px 75px]{{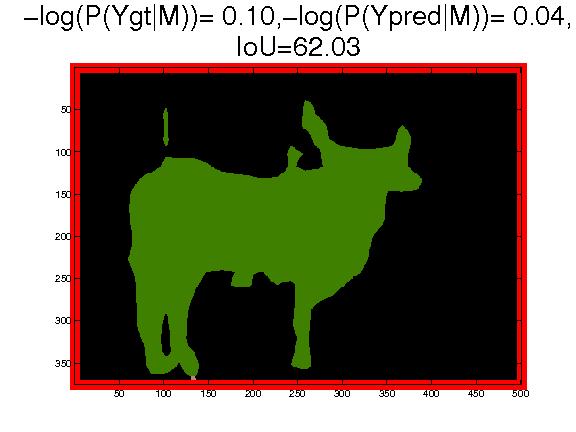}}}}}&
\res{\includegraphics[width=1.1in, height=0.8in,clip=true, trim= 72px 45px 40px 65px]{{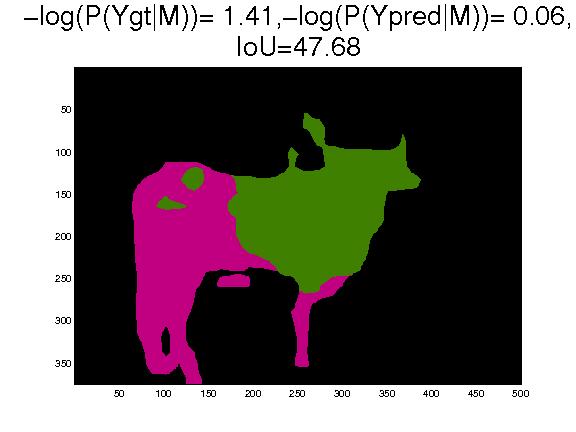}}}&
\res{\includegraphics[width=1.1in, height=0.8in,clip=true, trim= 72px 45px 40px 65px]{{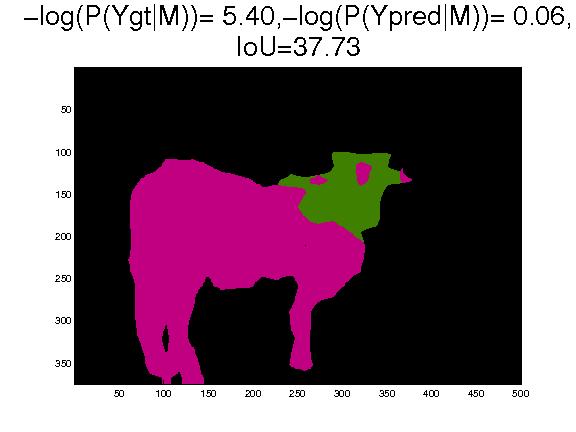}}}\\[-3px]
&  {\scriptsize IoU 54.26} &{\scriptsize IoU 56.45} & {\textbf{\scriptsize IoU 62.03}} & {\scriptsize IoU 47.68} & {\scriptsize IoU 37.73} \\[-7px]

\res{\includegraphics[width=1.1in,height=0.8in]{{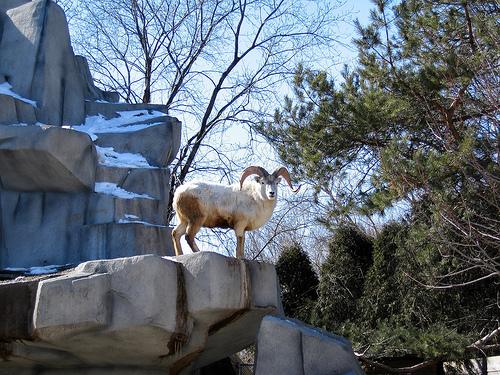}}} &
\res{\includegraphics[width=1.1in, height=0.8in, clip=true, trim= 80px 60px 70px 75px]{{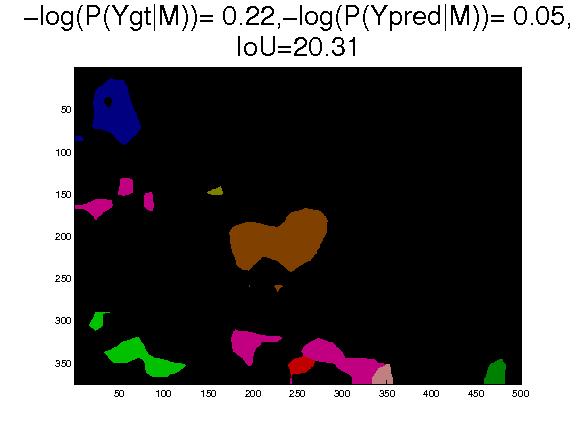}}}&
\res{\includegraphics[width=1.1in, height=0.8in,clip=true, trim= 72px 45px 40px 65px]{{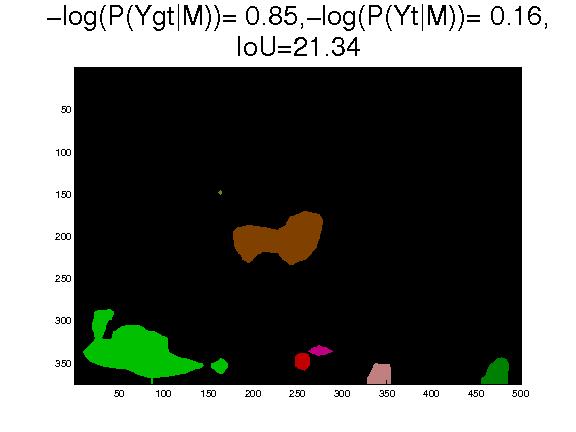}}} &
\res{\includegraphics[width=1.1in, height=0.8in,clip=true, trim= 72px 45px 40px 65px]{{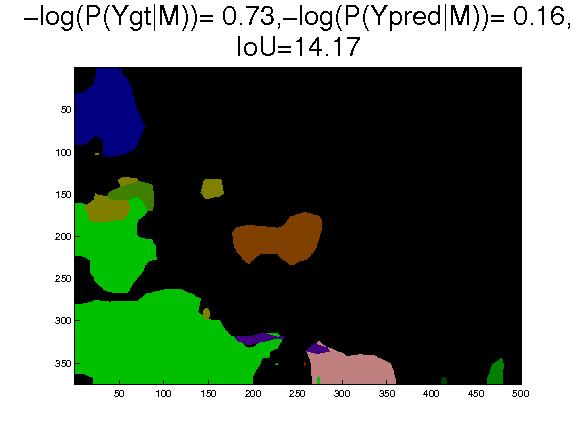}}} &
\raisebox{2pt}{\res{\fcolorbox{red}{white}{\includegraphics[width=1.1in, height=0.8in, clip=true, trim= 80px 60px 70px 75px]{{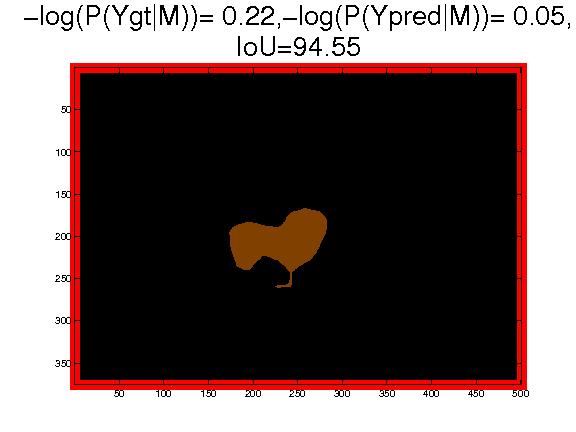}}}}}&
\res{\includegraphics[width=1.1in, height=0.8in,clip=true, trim= 72px 45px 40px 65px]{{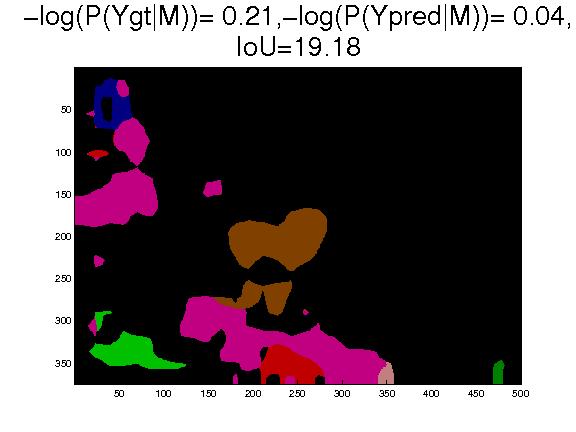}}}\\[-3px]
&  {\scriptsize IoU 20.31} & {\scriptsize IoU 21.34} & {\scriptsize IoU 14.17} & {\textbf{\scriptsize IoU 94.55}} & {\scriptsize IoU 19.18} \\

Input & & Net 1 & Net 2 &  Net 3 &  Net 4 \\[3px]

\end{tabular}
}
\caption{Samples images and corresponding predictions obtained by each member of the sMCL ensemble as well as the top output of a classical ensemble. The output with minimum loss on each example is outlined in red. Notice that sMCL ensembles vary in the shape, class, and frequency of predicted segments.}
\label{fig:segqual}
\end{figure}

\subsection{Image Captioning}
\label{captioning}

In this section, we show that sMCL trained ensembles can produce sets of high quality and diverse sentences, which is essential to improving recall and capturing ambiguities in language and perception.

\begin{table}[b]
\centering
\resizebox{\columnwidth}{!}{
\begin{tabular}{c@{\hskip0.25in} c c c c c@{\hskip0.4in} c c c c c c}
& \multicolumn{5}{C{2.2in}}{Oracle CIDEr-D for Ensemble of Size} & \multicolumn{5}{c}{\# Unique n-Grams (M=5)}\\
 & M = 1 & 2 & 3 & 4 & 5 & & n = 1 & 2 & 3 & 4 & \multicolumn{1}{C{0.75in}}{Avg.\newline Length}\\
 \hline\\[-4pt]
sMCL                     & -     & 0.822 & 0.862 & 0.911 & 0.922 & & 713 & 2902 & 6464 & 15427  & 10.21\\
MCL~\cite{rivera_nips12} & -     & 0.752 & 0.81 & 0.823 & 0.852 & & 384 & 1565 & 3586 & 9551 & 9.87\\
Dey~\cite{Dey_2015_ICCV} & -     & 0.798 & 0.850 & 0.887 & 0.910 & & 584 & 2266 & 4969 & 12208 & 10.26\\
Indp.                    & 0.684 & 0.757 & 0.784 & 0.809 & 0.831 & & 540 & 2003 & 4312 & 10297 & 10.24\\[2pt]
\hline\\[-4pt]
sMCL (fine-tuned CNN)    & -     & 1.064  &1.130 & 1.179 & 1.184 & & 1135 & 6028 & 15184 & 35518 & 10.43\\
Indp. (fine-tuned CNN)   & 0.912 & 1.001 & 1.05  & 1.073 & 1.095 & & 921 & 4335 & 10534 & 23811 & 10.33\\[2pt]
\hline\\[-4pt]
Beam Search              & 0.654 & 0.754 & 0.833 & 0.888 & 0.943 & & 580 & 2272 & 4920 & 12920 & 10.62\\[2pt]
\end{tabular}}
\caption{sMCL base methods outperform other ensemble methods a captioning, improve both oracle performance and the number of distinct n-grams. For low M, sMCL also performs better than multiple-output decoders.}
\label{tab:caption}
\end{table}

\xhdr{Model.} We adopt the model and training procedure of Karpathy et al.~\cite{karpathy2015deep}, utilizing their publicly available implementation \emph{neuraltalk2}. The model consists of an VGG16 network \cite{chatfield2014return} which encodes the input image as a fixed-length representation for a Long Short-Term Memory (LSTM) language model. We train and test on the MSCOCO dataset \cite{coco}, using the same splits as \cite{karpathy2015deep}. We perform two experimental setups by either freezing or finetuning the CNN. In the first, we freeze the parameters of the CNN and train multiple LSTM models using the CNN as a static feature generator. In the second, we aggregate and back-propagate the gradients from each LSTM model through the CNN in a tree-like model structure. This is largely a construct of memory restrictions as our hardware could not accommodate multiple VGG16 networks. We train each ensemble for 70k iterations with the parameters of the CNN fixed. For the fine-tuning experiments, we perform another 70k iterations of training to fine-tune the CNN. We generate sentences for testing by performing beam search with a beam width of two (following \cite{karpathy2015deep}).

\xhdr{Results.} Table \ref{tab:caption} presents the oracle CIDEr-D~\cite{Vedantam_2015_CVPR} scores for all methods on the validation set. We additionally compare with all outputs of a beam search over a single CNN+LSTM model with beam width ranging from 1 to 5. sMCL significantly outperforms the baseline ensemble learning methods (shown in the upper section of the table), increasing both oracle performance and the number of unique n-grams. For $M=5$, beam search from a single model achieves greater oracle but produces significantly fewer unique n-grams. We note that beam search is an inference method and increased beam width could provide similar benefits for sMCL ensembles.

\begin{figure}[t]
\centering
\resizebox{1.025\columnwidth}{!}{
\hspace{-10pt}
\begin{tabular}{p{1.2in} p{3in} p{3in}}
	\begin{center}Input\end{center} &  \begin{center}Independently Trained Networks\end{center} &  \begin{center}sMCL Ensemble \end{center} \\[-10px]
	\cmidrule(lr){1-1}\cmidrule(lr){2-2}\cmidrule(lr){3-3} \\[-5px]
	
	\centering\raisebox{-.45\height}{\includegraphics[width=1.2in, height=0.8in]{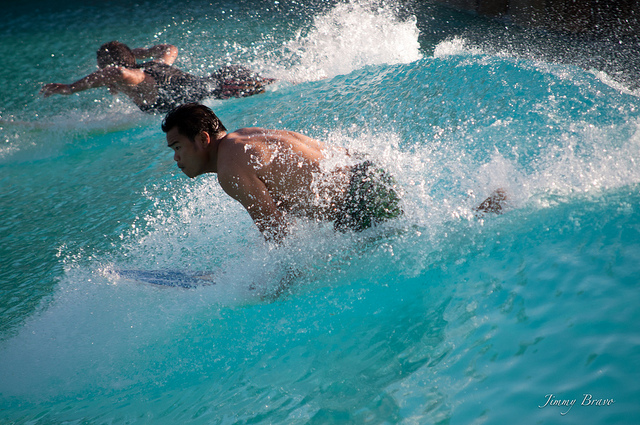}} & 
	\begin{tabular}{l}
	A man riding a wave on top of a surfboard.	\\
	A man riding a wave on top of a surfboard.	\\
	A man riding a wave on top of a surfboard.	\\
	A man riding a wave on top of a surfboard.
	\end{tabular}
	&
	\begin{tabular}{l}
	A man riding a wave on top of a surfboard.	\\
	A person on a surfboard in the water.	\\
	A surfer is riding a wave in the ocean.	\\
	A surfer riding a wave in the ocean.	
	\end{tabular}\\[0.35in]

	\centering\raisebox{-.45\height}{\includegraphics[width=1.2in, height=0.8in]{images/cap_examples/2.jpg}} & 
	\begin{tabular}{l}
	A group of people standing on a sidewalk.\\
	A man is standing in the middle of the street.\\	
	A group of people standing around a fire hydrant.\\	
	A group of people standing around a fire hydrant	
	\end{tabular}
	&
	\begin{tabular}{l}
	A man is walking down the street with an umbrell.\\
	A group of people sitting at a table with umbrellas.\\	
	A group of people standing around a large plane.\\	
	A group of people standing in front of a building	
	\end{tabular}\\[0.35in]

	\centering\raisebox{-.45\height}{\includegraphics[width=1.2in, height=0.8in]{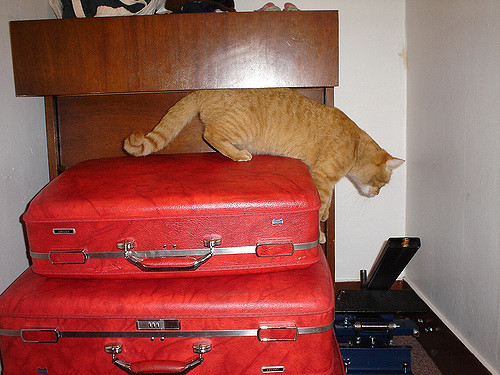}} & 
	\begin{tabular}{l}
	A kitchen with a stove and a microwave.\\
	A white refrigerator freezer sitting inside of a kitchen.\\
	A white refrigerator sitting next to a window.\\
	A white refrigerator freezer sitting in a kitchen	
	\end{tabular}
	&
	\begin{tabular}{l}
	A cat sitting on a chair in a living room.\\	
	A kitchen with a stove and a sink.\\
	A cat is sitting on top of a refrigerator.\\
	A cat sitting on top of a wooden table	
	\end{tabular}\\[0.35in]
	\centering\raisebox{-.45\height}{\includegraphics[width=1.2in, height=0.8in]{images/cap_examples/5.jpg}} & 
	\begin{tabular}{l}
	A bird is sitting on a tree branch.\\
	A bird is perched on a branch in a tree.\\	
	A bird is perched on a branch in a tree.\\	
	A bird is sitting on a tree branch		
	\end{tabular}
	&
	\begin{tabular}{l}
	A small bird perched on top of a tree branch.\\	
	A couple of birds that are standing in the grass.\\	
	A bird perched on top of a branch.\\	
	A bird perched on a tree branch in the sky	
	\end{tabular}
\end{tabular}
}
\vspace{5pt}
\caption{Comparison of sentences generated by members of a standard independently trained ensemble and an sMCL based ensemble of size four.}
\label{fig:captions}
\end{figure}

\xhdr{Intepretable Expertise: sMCL as N-Gram Specialists.} Figure \ref{fig:captions} shows example images and generated captions from standard and sMCL ensembles of size four (results from beam search over a single model are similar). It is evident that the independently trained models tend to predict similar sentences independent of initialization, perhaps owing to the highly structured nature of the output space and the mode bias of the underlying language model. On the other hand, the sMCL based ensemble generates diverse sentences which capture ambiguity both in language and perception. The first row shows an extreme case in which all of the members of the standard ensemble predict identical sentences. In contrast, the sMCL ensemble produces sentences that describe the scene with many different structures. In row three, both models are confused about the content of the image, mistaking the pile of suitcases as kitchen appliances. However, the sMCL ensemble widens the scope of some sentences to include the cat clearly depicted in the image. The fourth row is an example of regression towards the mode, with the standard model producing multiple similar sentences describing birds on branches. In the sMCL ensemble, we also see this tendency; however, one model breaks away and captures the true content of the image.

\section{Conclusion}
\label{sec:conclusions}
To summarize, we propose Stochastic Multiple Choice Learning (sMCL), an SGD-based technique for training diverse deep ensembles that follows a `winner-take-gradient' training strategy. Our experiments demonstrate the broad applicability and efficacy of sMCL for training diverse deep ensembles. In all experimental settings, sMCL significantly outperforms classical ensembles and other strong baselines including the 5x slower MCL procedure. Our analysis shows that exactly the same algorithm (sMCL) automatically generates specializations among ensemble members along different task-specific dimensions. sMCL is simple to implement, agnostic to both architecture and loss function, parameter free, and simply involves introducing one new sMCL layer into existing ensemble architectures.

\subsubsection*{Acknowledgments}
\smaller
This work was supported in part by a National Science Foundation
CAREER award, an Army Research Office YIP award, ICTAS Junior Faculty
award, Office of Naval Research grant N00014-14-1-0679, Google Faculty
Research award, AWS in Education Research grant, and NVIDIA GPU
donation, all awarded to DB, and by an NSF CAREER award (IIS-1253549),
the Intelligence Advanced Research Projects Activity (IARPA) via Air
Force Research Laboratory contract FA8650-12-C-7212, a Google Faculty
Research award, and an NVIDIA GPU donation, all awarded to DC.
Computing resources used by this work are supported in part by NSF
(ACI-0910812 and CNS-0521433), the Lily Endowment, Inc., and the
Indiana METACyt Initiative. The U.S. Government is authorized to
reproduce and distribute reprints for Governmental purposes
notwithstanding any copyright annotation thereon. Disclaimer: The
views and conclusions contained herein are those of the authors and
should not be interpreted as necessarily representing the official
policies or endorsements, either expressed or implied, of IARPA, AFRL,
NSF, or the U.S. Government.

{\small
\bibliographystyle{ieee}
\bibliography{dbatra}
}

\end{document}